\documentclass[11pt,a4paper]{article}
\usepackage[margin=1in]{geometry}
\usepackage[T1]{fontenc}
\usepackage{amsmath}
\usepackage{newtxtext,newtxmath}
\usepackage{bm}
\usepackage{microtype}
\usepackage{authblk}
\usepackage{cite}
\usepackage[font=small,labelfont=bf]{caption}
\usepackage{graphicx}
\usepackage{booktabs}
\usepackage{array}
\usepackage{tabularx}

\newcolumntype{Y}{>{\centering\arraybackslash}X}
\usepackage{multirow}
\usepackage[table]{xcolor}
\usepackage{url}
\usepackage{placeins}
\usepackage[hidelinks]{hyperref}
\graphicspath{{figures/}}

\newcommand{\Fscd}{\ensuremath{F_{scd}}}
\newcommand{\SeK}{\ensuremath{\mathrm{SeK}}}
\newcommand{\BFP}{BFP}
\newcommand{\ie}{i.e.}
\newcommand{\eg}{e.g.}
\newcommand{\M}{\mathcal{M}}
\newcommand{\mF}{\mathbf{F}}
\newcommand{\mC}{\mathbf{C}}
\newcommand{\mP}{\mathbf{P}}
\newcommand{\mq}{\mathbf{q}}
\newcommand{\nul}{\varnothing}

\begin{document}

\title{MaskSCD: Bitemporal Mask Classification for Semantic Change Detection in Remote Sensing Images}

\author[1,2]{Weikang Yu}
\author[1,3]{Pedram Ghamisi}
\affil[1]{Helmholtz Institute Freiberg for Resource Technology, Helmholtz-Zentrum Dresden-Rossendorf, 09599 Freiberg, Germany}
\affil[2]{Chair of Data Science in Earth Observation, Technical University of Munich, Germany}
\affil[3]{Lancaster University, LA1 4YR Lancaster, U.K.}
\affil[ ]{\texttt{\{w.yu, p.ghamisi\}@hzdr.de}}
\date{}

\maketitle

\begin{abstract}
Semantic change detection (SCD) identifies where the land surface has changed and which land-cover transition has taken place. Most deep SCD networks predict the change map and the land-cover map of each date with separate per-pixel decoders. These outputs need not agree, so a pixel can be labeled as changed although it has the same class at both dates, and per-pixel decisions fragment objects. We present MaskSCD, which formulates SCD as the mask classification of a bitemporal label-pair image. Each bitemporal segment, a region of constant land-cover class pair, is predicted as one mask shared by both dates together with one class per date, and the change map is derived from the predicted class pair. This formulation keeps the change and semantic maps consistent and the label space linear in the number of classes. A bitemporal masked-attention transformer decoder implements it by reading each segment in both images and relating the two readings, and a bitemporal fusion pyramid provides per-date and change-aware features. MaskSCD is evaluated on SECOND, Landsat-SCD and Hi-UCD, which span 0.1 to 30\,m resolution and four to nine land-cover classes, against fourteen recent SCD models re-trained under one protocol. MaskSCD achieves the best separated kappa, semantic change $F_1$ and transition $F_1$ on all three datasets and exceeds the strongest competitor by about five points in mean separated kappa. It produces no contradictory change pixels, which account for nearly 30\,\% of the predicted change of some multi-task networks. The code is available at \url{https://github.com/EricYu97/MaskSCD}.
\end{abstract}


\section{Introduction}
Change detection (CD) identifies the surface changes between co-registered images of the same scene acquired at different times. It is a core tool for urban management, disaster assessment and land-use monitoring~\cite{wen2021vhrcd,cheng2024cdreview}, as well as for tracking the footprint of mining activities~\cite{yu2024minenetcd}. Binary change detection (BCD) only answers whether a pixel has changed. Semantic change detection (SCD) also recovers the land-cover class before and after the change, that is, the transition of every changed pixel~\cite{daudt2019hrscd}. Many applications need this ``from--to'' information, for example the conversion of farmland to buildings or roads in urban planning and the drainage of wetland for agriculture in hydrology. Benchmarks with dense bitemporal land-cover annotation now exist for aerial~\cite{yang2021asymmetric,tian2022hiucd}, high-resolution satellite~\cite{tan2025triples} and Landsat imagery~\cite{yuan2022landsatscd}, and a large number of deep SCD networks have been developed on them.

\begin{figure}[t]
\centering
\includegraphics[width=0.5\textwidth]{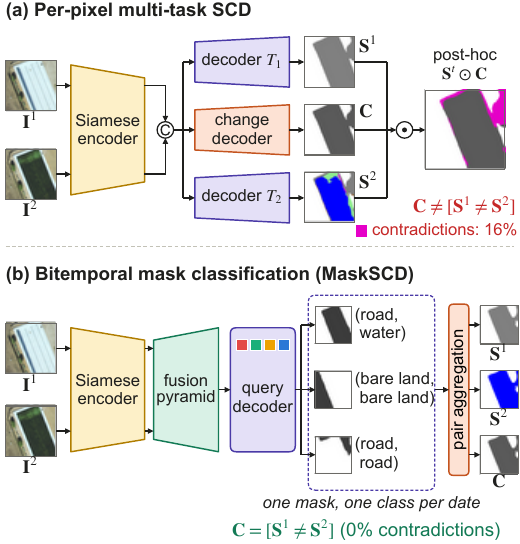}
\caption{Two paradigms for SCD on a Hi-UCD sample. (a) Per-pixel multi-task SCD; magenta: contradictory change pixels of Mamba-FCS~\cite{wijenayake2025mambafcs} (16 percent of its predicted change). (b) Bitemporal mask classification (MaskSCD).}
\label{fig:paradigm}
\end{figure}

Nearly all deep SCD networks share one design, which we refer to as per-pixel multi-task SCD. A Siamese encoder extracts bitemporal features, three per-pixel decoders produce a binary change map and one semantic map per date, and the semantic maps are then masked with the change map (Fig.~\ref{fig:paradigm}(a))~\cite{daudt2019hrscd,ding2022bisrnet}. Progress within this paradigm has come from richer interaction between the branches, for example explicit transition modeling~\cite{ding2024scannet}, foreground and background co-guidance~\cite{zhang2025foba} and semantic consistency losses~\cite{wang2024cdsc}, and from stronger encoders, most recently state-space models~\cite{chen2024changemamba} and vision foundation models~\cite{tong2026semdino}. The paradigm itself, however, has two weaknesses.

First, change and semantics are decided independently at every pixel, so the three maps are not guaranteed to agree. A pixel can be declared changed while receiving the same class at both dates, or unchanged while receiving two different classes. We call the first case a contradictory change pixel. In our benchmark of fourteen recent models on three datasets, every model produces such pixels, and on the full-coverage Hi-UCD dataset they make up nearly 30 percent of the predicted change for some models. An auxiliary consistency loss can reduce the number of contradictory pixels but cannot remove them. Inconsistent maps are difficult to use in decision making, and reliability is a central requirement of responsible artificial intelligence for Earth observation~\cite{ghamisi2025responsible}. Second, per-pixel classification inherits the known defects of pixel-wise labeling: fragmented object boundaries, speckle noise and incomplete delineation of homogeneous objects. Our earlier work MaskCD~\cite{yu2024maskcd} showed that these defects can be overcome for BCD by generating and classifying object masks, following the mask classification paradigm of MaskFormer and Mask2Former~\cite{cheng2021maskformer,cheng2022mask2former}.

Extending mask classification from BCD to SCD raises two problems. The first is the definition of a segment. In BCD a segment is a changed object. In SCD every pixel carries a pair of labels, and a na\"ive extension would assign each mask to one of the possible transitions, that is, to an ordered pair of land-cover classes. This option is known to suffer from the extreme imbalance of transition types~\cite{daudt2019hrscd}, and its label space grows quadratically with the number of land-cover classes. Alternatively, each date can be segmented with its own masks, as in DMCTNet~\cite{yang2025dmctnet}, but then the change map has to come from an additional per-pixel branch, and change and semantics are again decided separately. The second is the decoding of a segment that spans two dates. A query of MaskCD~\cite{yu2024maskcd} has a single state that reads the concatenated features of the two dates. It sees the dates only in mixed form and cannot tell which class its region takes at which date. Recovering the class pair of a region requires reading it once in each image and comparing the two readings.

We address both issues with MaskSCD (Fig.~\ref{fig:paradigm}(b)). Its central idea is to regard SCD as the segmentation of the \emph{bitemporal label-pair image}. A \emph{bitemporal segment} is a region in which the pair of land-cover classes at the two dates is constant, and unchanged land is the segment whose two classes are the same. Each query localizes one bitemporal segment through a single mask shared by both dates and classifies it once per date with a common classifier. Because the change map follows from the predicted pair, change and semantics always agree, and the label space grows linearly with the number of classes. To decode bitemporal segments, we propose a bitemporal masked-attention transformer decoder. Each query carries two temporal states that share one content embedding. In each layer, the two states cross-attend to their own dates within the shared mask of the query and read the same region in both images. A joint self-attention over all states then lets the two states of a query compare their readings and lets different queries compete for regions. The decoder is fed by a bitemporal fusion pyramid (\BFP), which fuses the dates at each pyramid level with bidirectional cross-temporal attention, keeps per-date features for the two states and adds a deeply supervised change pyramid to the shared mask feature.

We evaluate MaskSCD on three SCD benchmarks, SECOND~\cite{yang2021asymmetric}, Landsat-SCD~\cite{yuan2022landsatscd} and Hi-UCD~\cite{tian2022hiucd}, which contain 0.5 to 3\,m aerial, 30\,m Landsat and 0.1\,m aerial imagery, cover both annotation conventions of SCD and distinguish four to nine land-cover classes. We compare MaskSCD with fourteen representative SCD models, from CNN multi-task networks to state-space, foundation-model-based and query-based methods, all re-trained in one code base under an identical protocol. The main contributions of this article are summarized as follows.
\begin{itemize}
\item We propose MaskSCD, which formulates SCD as the mask classification of the bitemporal label-pair image. Each query carries one mask shared by both dates and one class label per date, a formulation that, to our knowledge, has not been used for SCD before. The change map is derived from the predicted class pair. It always agrees with the semantic maps, and the label space grows only linearly with the number of classes.
\item We design a bitemporal masked-attention transformer decoder for this formulation. Each query keeps two temporal states, which attend to their own date within the shared mask, compare their readings through joint bitemporal self-attention and are labeled by one classifier shared across dates. A bitemporal fusion pyramid with cross-temporal attention and a deeply supervised change pyramid provides the per-date memories and the change-aware mask feature of the decoder.
\item We conduct a unified benchmark of fourteen recent SCD models on three datasets, all re-trained under one protocol, and complement the standard metrics with a pixel-weighted F1 score of fully correct transitions and a contradiction measure. MaskSCD achieves the best separated kappa, semantic change F1 score and F1 score of fully correct transitions on every dataset and produces no contradictory change pixels.
\end{itemize}

The rest of this article is organized as follows. Section~\ref{sec:related} reviews related work. Section~\ref{sec:method} presents MaskSCD. Section~\ref{sec:exp} reports the experimental results and ablation studies, and Section~\ref{sec:conclusion} concludes the article.

\section{Related Work}
\label{sec:related}

\subsection{Semantic Change Detection}
SCD extends binary change detection with the land-cover classes before and after the change. Daudt et al.~\cite{daudt2019hrscd} distinguished three strategies for it. Post-classification comparison classifies the two dates independently and compares the resulting land-cover maps~\cite{wen2021vhrcd}. It needs no change labels, but the classification errors of both dates accumulate in the change map. Direct classification assigns every pixel to one of the $K^2$ transitions between $K$ classes. Rare transitions then have very few training samples, and the label space grows quadratically. Multi-task learning shares one encoder among a binary change head and two land-cover heads. It outperformed both alternatives~\cite{daudt2019hrscd} and has dominated the field since.

Multi-task networks differ mainly in how their branches exchange information. Yang et al.~\cite{yang2021asymmetric} introduced asymmetric Siamese networks together with the SECOND dataset and the separated kappa (\SeK). Bi-SRNet~\cite{ding2022bisrnet} merges the semantic features of both dates in its change branch, reasons over single-date and cross-date correlations with semantic reasoning blocks and introduced the semantic change $F_1$ score (\Fscd). ChangeMask~\cite{zheng2022changemask} decouples SCD into temporal-wise semantic segmentation and binary change detection, with a semantic-aware encoder and a temporal-symmetric transformer for the change representation. SCanNet~\cite{ding2024scannet} models the ``from--to'' transitions explicitly with a semantic change transformer. LSAFNet~\cite{zhou2024lsafnet} keeps the semantic branches of the two dates apart until a late-stage change decoder fuses their decoded features. HGINet~\cite{long2024hginet} models the interaction between the bitemporal semantic features and their differences with a graph convolutional network, and CPGNet~\cite{long2025cpgnet} adds an auxiliary change branch whose output serves as a change prior. TripleS~\cite{tan2025triples} addresses the conflicts between the subtasks during training with stepwise optimization and selective parameter binding.

Since the three maps come from separate heads, they can contradict each other, and a second group of networks targets the agreement between the outputs. The semantic consistency loss of Bi-SRNet~\cite{ding2022bisrnet} pulls the bitemporal semantic features of unchanged pixels together and pushes those of changed pixels apart, and later networks such as LSAFNet~\cite{zhou2024lsafnet} adopt it. CdSC~\cite{wang2024cdsc} aligns the semantics of the two dates with a contrastive co-alignment loss, and FoBa~\cite{zhang2025foba} combines foreground and background guidance with a consistency loss. ClearSCD~\cite{tang2024clearscd} infers the binary change from the land-cover posteriors of the two dates. These networks encourage agreement through a loss or by comparing posteriors, but they still decide change and semantics pixel by pixel and inherit the fragmented objects and speckle of per-pixel labeling.

Other recent networks adopt stronger encoders. ChangeMamba~\cite{chen2024changemamba} builds on the VMamba encoder~\cite{liu2024vmamba} and learns spatio-temporal relations in its change decoder by scanning the tokens of the two dates sequentially, interleaved and stacked along the channels. FoBa and Mamba-FCS~\cite{wijenayake2025mambafcs} use the same encoder, the latter with spatio-frequency fusion, change-guided attention and a loss inspired by \SeK. GSTM-SCD~\cite{liu2025gstmscd} adds a spatio-temporal graph to the state-space model to strengthen the interaction between the dates. Vision foundation models provide the encoder of PerASCD~\cite{shen2026perascd}, which builds on the PerA remote sensing foundation model, and of SemDINO~\cite{tong2026semdino}, which fuses DINOv3 features with convolutional details.

Query-based decoders have only recently been applied to SCD. DMCTNet~\cite{yang2025dmctnet} feeds multi-scale features to a masked query decoder to improve the completeness of changed objects. It segments each date with its own masks and gates the resulting land-cover maps with a separate per-pixel change map, which leaves change and semantics to different outputs. In MaskSCD, one mask and one class pair per bitemporal segment determine both the semantic maps and the change map.

\subsection{Bitemporal Feature Fusion}
Change detection methods differ in where and how they combine the two dates. Classical methods compare the dates directly in the image domain. Image differencing, ratioing and change vector analysis~\cite{wen2021vhrcd} produce a difference image that is then thresholded, and object-based methods compare image segments to suppress salt-and-pepper noise~\cite{cheng2024cdreview}. These comparisons rely on hand-crafted features and thresholds that transfer poorly across sensors and scenes.

Deep networks learn this comparison from data. Daudt et al.~\cite{daudt2018fcsiam} compared early fusion, which concatenates the images at the input, with Siamese networks that fuse the features of the two dates by concatenation or differencing in the decoder. Subsequent CNNs fuse and supervise the features at several levels, as DSIFN~\cite{zhang2020dsifn} does with attention and deep supervision on multi-level difference features. Attention-based detectors let each date attend to the other. BIT~\cite{chen2021bit} models the bitemporal context with a compact set of semantic tokens, and DMINet~\cite{feng2023dminet} lets the dates interact through joint self- and cross-attention before their difference is computed. Changer~\cite{fang2023changer} moves the interaction into the encoder and shows that simple spatial and channel exchange between the two feature streams works well.

More recent detectors build the fusion around large pre-trained encoders, for example through the bitemporal adapter branch that BAN~\cite{li2024ban} attaches to a frozen foundation model. Explicit difference features, obtained by subtraction or concatenation, remain the most common change signal~\cite{zhang2020dsifn,feng2023dminet}, and in multi-task SCD networks the dates are usually fused only inside the change branch~\cite{ding2022bisrnet,zhou2024lsafnet}.

Most of these modules end in a single fused change representation that feeds a per-pixel head, and MaskCD~\cite{yu2024maskcd} likewise forms its change representation once from the concatenated backbone features of the two dates. A query decoder with two temporal states needs per-date features from which each state can read the semantics of its own date, and change-aware features at each scale of its multi-scale memory. The proposed fusion pyramid provides both, as it fuses the dates at each level but keeps their fused features separate and supervises an explicit change feature at each scale.

\subsection{Mask Classification}
Mask classification splits segmentation into predicting a set of binary masks and assigning a class to each mask. DETR~\cite{carion2020detr} introduced the underlying set prediction with learnable queries and bipartite matching for object detection. MaskFormer~\cite{cheng2021maskformer} carried it over to segmentation and showed that semantic and panoptic segmentation can both be solved by predicting class-labeled masks. Mask2Former~\cite{cheng2022mask2former} introduced masked attention, which restricts the cross-attention of each query to the foreground of its predicted mask, and became a universal architecture for semantic, instance and panoptic segmentation. Because these models predict whole segments, they yield coherent objects with sharp boundaries.

Query-based models have also been extended across time. In video instance segmentation, one query of Mask2Former predicts a spatio-temporal mask tube shared across frames~\cite{cheng2021mask2formervis}. Such a query follows one region through time, but the model is designed to track object identity over many frames and does not describe how the class of a region changes between two dates.

In remote sensing, mask classification has mostly served binary change detection. MaskChanger~\cite{ebrahimzadeh2024maskchanger} and MaskCD~\cite{yu2024maskcd} reformulated change detection as the classification of change masks, the latter with a deformable cross-level change representation and a masked-attention decoder. CDMaskFormer~\cite{ma2024cdmaskformer} adapts its change queries to the content of the bitemporal images, NeXt2Former-CD~\cite{wang2026next2former} combines a DINOv3-initialized ConvNeXt encoder with deformable temporal fusion and a Mask2Former decoder, and TextMCD~\cite{wu2026textmcd} supervises the category of each change mask with text labels. All of these networks output binary change masks, and what a mask should represent when every pixel carries a pair of labels has not been addressed. MaskSCD addresses it with the bitemporal segment. As in video segmentation, one query follows one segment through time, but its two temporal states receive one class each, and the change label follows from the pair of classes.

\section{Methodology}
\label{sec:method}
\begin{figure*}[t]
\centering
\includegraphics[width=\textwidth]{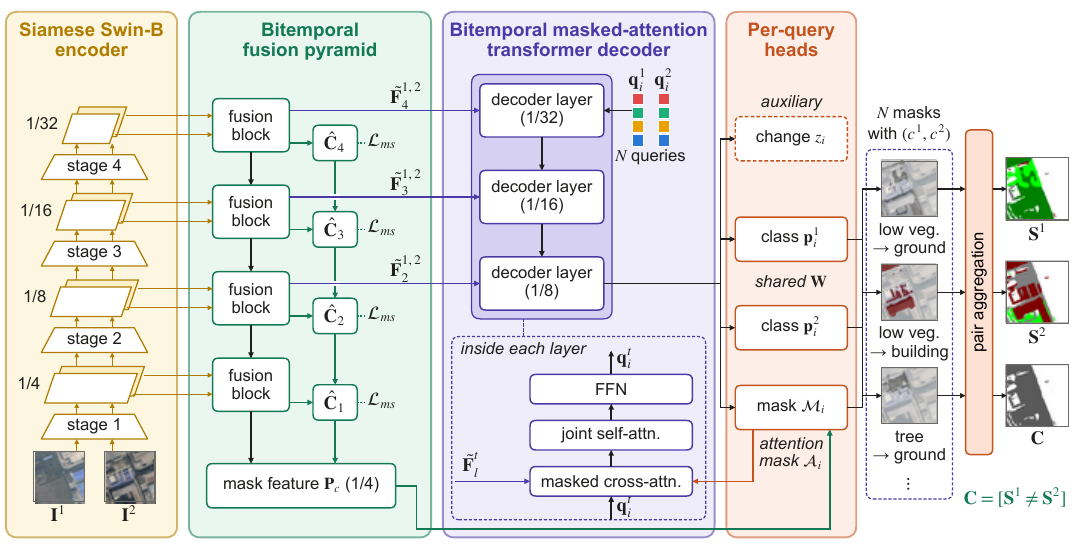}
\caption{Overall framework of MaskSCD. The fusion block is detailed in Fig.~\ref{fig:bfp}, the decoder layer and the heads in Fig.~\ref{fig:decoder}. The dashed change head is used only in training.}
\label{fig:overview}
\end{figure*}

This section first formulates SCD as the mask classification of a bitemporal label-pair image and then describes the components of MaskSCD (Fig.~\ref{fig:overview}): the Siamese encoder, the bitemporal fusion pyramid, the bitemporal masked-attention transformer decoder, the inference by pair aggregation and the training objective.

\subsection{Problem Formulation}
\label{sec:formulation}
Let $\mathbf{I}^1,\mathbf{I}^2\in\mathbb{R}^{3\times H\times W}$ be co-registered images with land-cover maps $\mathbf{L}^1,\mathbf{L}^2\in\{0,1,\dots,K{-}1\}^{H\times W}$ over $K$ classes. SCD predicts the semantic maps $\mathbf{S}^1,\mathbf{S}^2$ and the change map $\mC$. On datasets annotated only inside changed regions (SECOND and Landsat-SCD), class $0$ is the ``no-change'' class, which unchanged pixels carry at both dates. On full-coverage datasets (Hi-UCD), class $0$ marks unlabeled pixels, and every labeled pixel carries its land-cover class at both dates. In both cases, the ground-truth change map is $[\mathbf{L}^1\neq\mathbf{L}^2]$.

\begin{figure}[t]
\centering
\includegraphics[width=0.6\textwidth]{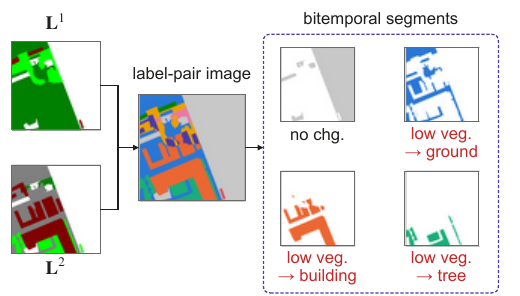}
\caption{Bitemporal segments of a SECOND sample, the regions of constant class pair in the label-pair image. Changed segments are labeled in red.}
\label{fig:labelpair}
\end{figure}

We view the pair $(\mathbf{L}^1,\mathbf{L}^2)$ as a single bitemporal label-pair image in which each pixel carries the pair $(c^1,c^2)$, and define a bitemporal segment as the region of one constant pair (Fig.~\ref{fig:labelpair}). Unchanged land is the segment $(c,c)$ (or $(0,0)$) and is treated like any other segment. SCD then becomes the mask classification of this image. A model outputs a set of $N$ predictions $\{o_i\}_{i=1}^N$ with $o_i=(\M_i,\mathbf{p}^1_i,\mathbf{p}^2_i)$, where $\M_i\in[0,1]^{H\times W}$ is a mask shared by both dates and $\mathbf{p}^t_i\in\Delta^{K+1}$ is a class distribution over the $K$ classes and a no-object class $\nul$ for date $t$. Both semantic maps are decoded from this set, and the change map follows as $\mC=[\mathbf{S}^1\neq\mathbf{S}^2]$. Compared with classifying each mask into one of $K^2$ transitions, the factorized formulation needs only $2(K{+}1)$ outputs per query with a classifier shared across dates, and each training segment supervises two $K$-way decisions. Rare transitions share their supervision with all segments that have the same class at either date.

\subsection{Overview}
As shown in Fig.~\ref{fig:overview}, a weight-shared hierarchical transformer encoder extracts four-level features $\{\mF^t_l\}_{l=1}^{4}$ at strides $4$, $8$, $16$ and $32$ from each image. The bitemporal fusion pyramid processes the two feature hierarchies from coarse to fine; at each level, a bitemporal fusion block fuses the two dates and emits per-date features $\tilde{\mF}^1_l,\tilde{\mF}^2_l$ and a change feature $\mC_l$. The per-date features at strides $32$, $16$ and $8$ form the memory of the bitemporal masked-attention transformer decoder, and the stride-4 features and the change pyramid form the shared mask feature $\mP_c$. The decoder refines $N$ queries with two temporal states each, and the heads predict one mask and two class distributions per query. Pair aggregation turns them into $\mathbf{S}^1$, $\mathbf{S}^2$ and $\mC$. Training uses bipartite matching between predictions and bitemporal segments.

\subsection{Hierarchical Transformer-Based Siamese Encoder}
\label{sec:encoder}
As in MaskCD~\cite{yu2024maskcd}, we adopt a Swin Transformer~\cite{liu2021swin} with shared weights as the Siamese encoder, here the base model (Swin-B) pre-trained on ImageNet-22k and fine-tuned on ImageNet-1k. Window-based and shifted-window multi-head self-attention alternate within each stage to build hierarchical context, and the shared weights embed both images into the same feature space. The four stage outputs $\mF^t_l\in\mathbb{R}^{D_l\times H_l\times W_l}$ ($l=1,\dots,4$) of each date are projected to $d=256$ channels by lateral $1\times1$ convolutions with group normalization and passed to the fusion pyramid. Unlike in MaskCD, the features of the two dates are kept separate at this point, and the fusion pyramid models their interaction at each level.

\begin{figure}[t]
\centering
\includegraphics[width=0.55\textwidth]{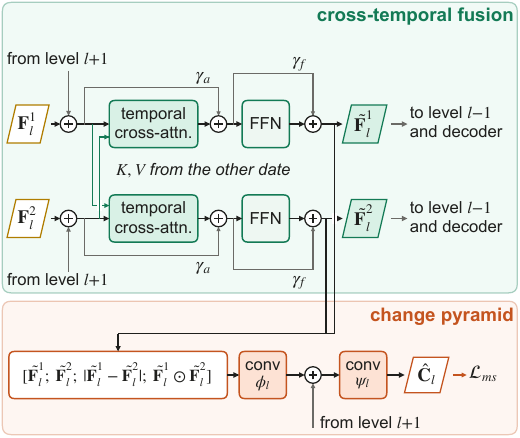}
\caption{Bitemporal fusion block at level $l$. Top: cross-temporal fusion with gated residuals $\gamma_a,\gamma_f$. Bottom: change feature and deeply supervised change pyramid.}
\label{fig:bfp}
\end{figure}

\subsection{Bitemporal Fusion Pyramid}
\label{sec:bfp}
The \BFP\ is a feature pyramid network (FPN)~\cite{lin2017fpn} whose lateral connections are extended by bitemporal fusion blocks (Fig.~\ref{fig:bfp}). Each fusion block consists of a cross-temporal fusion, which exchanges information between the dates through bidirectional cross-temporal attention and a gated convolutional FFN, and an explicit change feature, which feeds the change pyramid. The pyramid processes the levels from coarse to fine. At level $l$, the up-sampled output of level $l{+}1$ is added to the lateral projection of $\mF^t_l$ for each date, the sum is processed by the fusion block, and the fused features are refined by a $3\times3$ convolution with group normalization before they are passed to the next finer level and to the decoder. All weights of the pyramid are shared by the two dates, and only the temporal embeddings introduced below tell the dates apart.

\subsubsection{Bidirectional Cross-Temporal Attention}
Let $\mathbf{X}^1,\mathbf{X}^2\in\mathbb{R}^{H_lW_l\times d}$ be the flattened features of the two dates at level $l$, $\mathbf{E}\in\mathbb{R}^{H_lW_l\times d}$ the two-dimensional sine positional encoding, and $\bm{\tau}^1,\bm{\tau}^2\in\mathbb{R}^d$ learnable temporal embeddings. The update of date $1$ from date $2$ is
\begin{equation}
\begin{split}
\Delta\mathbf{X}^1=\mathrm{MHA}\big(&\mathrm{LN}(\mathbf{X}^1)+\mathbf{E}+\bm{\tau}^1,\\
&\mathrm{LN}(\mathbf{X}^2)+\mathbf{E}+\bm{\tau}^2,\ \mathrm{LN}(\mathbf{X}^2)\big),
\end{split}
\label{eq:xattn}
\end{equation}
where $\mathrm{LN}$ denotes layer normalization and $\mathrm{MHA}(Q,K,V)$ multi-head attention with 8 heads, whose query, key and value projections are shared by both directions; $\Delta\mathbf{X}^2$ is obtained symmetrically by exchanging the roles of the two dates. The shared positional encoding lets the attention relate tokens at the same location, and the temporal embedding tells the shared projections which date they are processing. The attention is global when a level has at most $4\,096$ tokens per date, which holds at strides $32$, $16$ and $8$ for $512\times512$ inputs and at every level for $256\times256$ inputs. At the stride-4 level of larger inputs, it is computed within non-overlapping $16\times16$ windows with padded keys masked out, which keeps the cost linear in the number of pixels.

\subsubsection{Gated Residuals and Convolutional FFN}
The attention outputs are added through learnable scalar gates that are initialized at zero,
\begin{equation}
\mathbf{X}^t\leftarrow\mathbf{X}^t+\gamma_a\Delta\mathbf{X}^t,\qquad
\mathbf{X}^t\leftarrow\mathbf{X}^t+\gamma_f\,\mathrm{FFN}(\mathrm{GN}(\mathbf{X}^t)),
\label{eq:gate}
\end{equation}
where $\mathrm{GN}$ denotes group normalization and the feed-forward network (FFN) consists of a $1\times1$ expansion to $2d$ channels, a $3\times3$ depth-wise convolution and a $1\times1$ projection with GELU activations. Zero initialization makes the block an identity mapping at the start of training, so the pyramid begins as a plain Siamese FPN and learns how much cross-temporal information to inject at each level.

\subsubsection{Explicit Change Features and Change Pyramid}
The fused per-date features $\tilde{\mF}^1_l,\tilde{\mF}^2_l$ are complemented by an explicit change feature that combines them with two symmetric statistics of the pair,
\begin{equation}
\mC_l=\phi_l\big([\tilde{\mF}^1_l;\ \tilde{\mF}^2_l;\ |\tilde{\mF}^1_l-\tilde{\mF}^2_l|;\ \tilde{\mF}^1_l\odot\tilde{\mF}^2_l]\big),
\label{eq:change}
\end{equation}
where $\phi_l$ is a $1\times1$ followed by a $3\times3$ convolution with group normalization. The absolute difference responds to any change of appearance, while the element-wise product measures agreement between the dates. The change features of the four levels are merged into a top-down change pyramid,
\begin{equation}
\hat{\mC}_l=\psi_l\big(\mC_l+\mathrm{up}(\hat{\mC}_{l+1})\big),\qquad \hat{\mC}_4=\psi_4(\mC_4),
\label{eq:pyramid}
\end{equation}
where $\psi_l$ is a $3\times3$ convolution with group normalization and $\mathrm{up}(\cdot)$ denotes bilinear up-sampling, so that coarse, context-rich change evidence guides the fine levels. A $1\times1$ binary change head is attached to each $\hat{\mC}_l$ and supervised with the ground-truth change map, which makes all levels of the pyramid encode change.

\subsubsection{Decoder Memory and Shared Mask Feature}
After the $3\times3$ refinement, the fused features at strides $32$, $16$ and $8$ are the per-date memories of the transformer decoder. The $1\times1$ projections $\mP^1,\mP^2$ of the stride-4 features and the finest change level form the shared mask feature
\begin{equation}
\mP_c=\mathrm{ChangeFusion}\big([\mP^1;\ \mP^2;\ |\mP^1-\mP^2|]\big)+\mathbf{W}_c\hat{\mC}_1,
\label{eq:maskfeat}
\end{equation}
where ChangeFusion consists of two $3\times3$ and one $1\times1$ convolution and $\mathbf{W}_c$ is a $1\times1$ convolution. Because a bitemporal segment occupies the same place in both images, a single mask feature suffices for both dates. $\mP_c$ is further up-sampled to stride $2$, added to the output of a light convolutional stem on $[\mathbf{I}^1;\mathbf{I}^2]$ and refined by a $3\times3$ and a $1\times1$ convolution, which sharpens the mask boundaries at a cost of $0.7$\,M parameters. Two auxiliary dense heads, a $K$-way semantic head on $\mP^1$ and $\mP^2$ and a binary change head on $\mP_c$, regularize the pixel features during training.

\begin{figure*}[t]
\centering
\includegraphics[width=\textwidth]{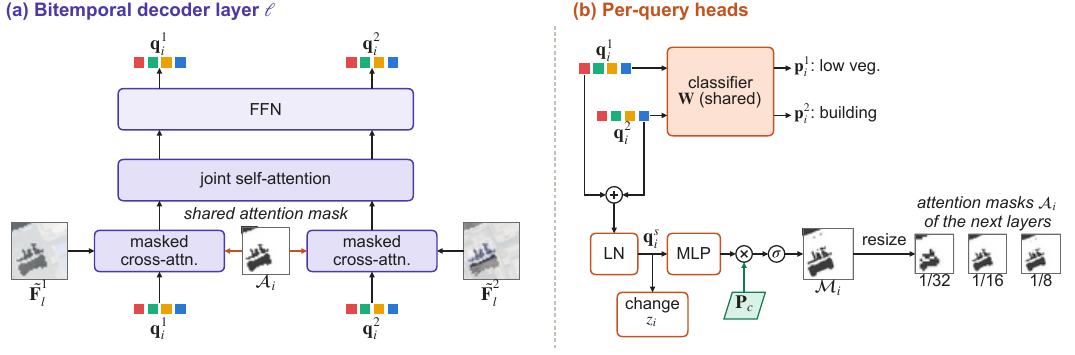}
\caption{Bitemporal masked-attention transformer decoder. (a) One decoder layer; the thumbnails show at stride 16 the region seen by both states of a query predicting ``low vegetation $\rightarrow$ building''. Residual connections and normalization are omitted. (b) Per-query heads.}
\label{fig:decoder}
\end{figure*}

\subsection{Bitemporal Masked-Attention Transformer Decoder}
\label{sec:decoder}
The decoder (Fig.~\ref{fig:decoder}) reads a bitemporal segment once in each image and relates the two readings. It follows the masked-attention design of Mask2Former~\cite{cheng2022mask2former} and MaskCD~\cite{yu2024maskcd}, but each query owns two temporal states. Query $i$ is initialized as $\mq^1_i=\mathbf{e}_i+\bm{\tau}^1_q$ and $\mq^2_i=\mathbf{e}_i+\bm{\tau}^2_q$, where $\mathbf{e}_i$ is a learnable content embedding and $\bm{\tau}^t_q$ are learnable temporal embeddings of the decoder, which are also added to the query positional embeddings. The two states start from the same content, which ties them to one location, and can develop different semantics.

\subsubsection{Masked Cross-Attention per Date}
In decoder layer $\ell\in\{1,2,3\}$, state $t$ attends to the memory of date $t$ at level $l=5-\ell$, \ie, at strides $32$, $16$ and $8$ in turn,
\begin{equation}
\mq^t_i\leftarrow\mathrm{LN}\Big(\mq^t_i+\mathrm{softmax}\big(\mathcal{A}_i+Q^t_iK_t^{\top}\big)V_t\Big),
\label{eq:maskedattn}
\end{equation}
where $Q^t_i$ is a linear projection of $\mq^t_i$, $K_t,V_t$ are linear projections of the memory of date $t$ plus a level embedding, positional embeddings are added to queries and keys, and $\mathcal{A}_i$ is $0$ inside and $-\infty$ outside the mask predicted for query $i$ by the previous layer, bilinearly resized to the resolution of level $l$ and binarized at $0.5$. The projection weights and the attention mask are shared by the two dates, so the two temporal states of a query read the same region, one in each image, while the mask suppresses the surrounding context. A query whose mask is empty attends to the whole image.

\subsubsection{Joint Bitemporal Self-Attention}
The $2N$ temporal states of all queries are then concatenated and processed by one self-attention layer,
\begin{equation}
[\mq^1_1,\dots,\mq^1_N,\mq^2_1,\dots,\mq^2_N]\leftarrow\mathrm{SelfAttn}\big([\mq^1_1,\dots,\mq^2_N]\big),
\label{eq:selfattn}
\end{equation}
followed by a shared FFN. The joint self-attention lets the two states of a query compare what they have seen at the two dates and recognize a change of their region, and it lets different queries compete for regions as in a standard set-prediction decoder. The decoder has $3$ layers, one per memory level, with 8 attention heads and an FFN width of $2\,048$, and uses $N=100$ queries with $d=256$.

\subsubsection{Heads}
After each layer, and once before the first layer to obtain the initial attention mask, the two states are fused into a shared localization state $\mq^s_i=\mathrm{LN}(\mq^1_i+\mq^2_i)$. A three-layer MLP maps it to a mask embedding, which is multiplied with the mask feature: $\M_i=\sigma\langle\mathrm{MLP}(\mq^s_i),\mP_c\rangle$. The class distributions $\mathbf{p}^1_i$ and $\mathbf{p}^2_i$ are obtained from the two temporal states with one linear classifier over $K{+}1$ classes shared across dates, and an auxiliary linear change logit $z_i=\mathbf{w}^\top\mq^s_i$ is supervised with the change flag of the matched segment. The shared classifier ties the semantics of the two dates to a common label space, in which ``water'' at date $1$ and ``water'' at date $2$ are the same output, and it lets unchanged segments supervise both states with the same target.

\subsection{Mask Classification of Label Pairs and Inference}
\label{sec:inference}
At inference, the $N$ predictions are aggregated per pixel into a score for every class pair,
\begin{equation}
s(c^1,c^2\,|\,x)=\sum_{i=1}^{N}\mathbf{p}^1_i(c^1)\,\mathbf{p}^2_i(c^2)\,\M_i(x),
\label{eq:pair}
\end{equation}
and the label pair of pixel $x$ is $(\mathbf{S}^1(x),\mathbf{S}^2(x))=\arg\max_{(c^1,c^2)}s(c^1,c^2|x)$ over the $K^2$ pairs, with the no-object probabilities dropped. The change map is $\mC=[\mathbf{S}^1\neq\mathbf{S}^2]$. This pair aggregation is the bitemporal counterpart of the semantic inference of MaskFormer~\cite{cheng2021maskformer}. It selects the class pair with the highest aggregated score and never produces a changed pixel with identical classes or an unchanged pixel with different classes. Test-time augmentation averages the score maps of (\ref{eq:pair}) over the eight dihedral transforms before the $\arg\max$.

\subsection{Optimization}
\label{sec:loss}
\subsubsection{Targets and Bipartite Matching}
The training targets of an image pair are its bitemporal segments: $o^{gt}_j=(\mathbf{m}^{gt}_j,c^1_j,c^2_j)$, one binary mask $\mathbf{m}^{gt}_j$ per distinct class pair $(c^1_j,c^2_j)$ present in the two label maps, together with the change flag $y_j=[c^1_j\neq c^2_j]$ (Fig.~\ref{fig:labelpair}). Predictions and targets are matched by the Hungarian algorithm~\cite{kuhn1955hungarian} with the cost
\begin{equation}
\begin{split}
\mathcal{L}_{match}(o_i,o^{gt}_j)=&-\tfrac{\lambda_{cls}}{2}\big(\mathbf{p}^1_i(c^1_j)+\mathbf{p}^2_i(c^2_j)\big)\\
&+\lambda_{m}\mathcal{L}_{mask}(\M_i,\mathbf{m}^{gt}_j),
\end{split}
\label{eq:match}
\end{equation}
where $\mathcal{L}_{mask}=\mathcal{L}_{bce}+\mathcal{L}_{dice}$~\cite{milletari2016dice} is evaluated on $12\,544$ uniformly sampled points~\cite{cheng2022mask2former}. Both dates enter the class term with equal weight, which favors queries that explain the classes of both dates.

\subsubsection{Loss Functions}
Given the optimal assignment $\hat\sigma$, the set loss of one decoder output is
\begin{equation}
\begin{split}
\mathcal{L}_{set}=\sum_{j}\Big[&\lambda_{cls}\big(\mathcal{L}_{ce}(\mathbf{p}^1_{\hat\sigma(j)},c^1_j)+\mathcal{L}_{ce}(\mathbf{p}^2_{\hat\sigma(j)},c^2_j)\big)\\
&+\lambda_{m}\mathcal{L}_{mask}(\M_{\hat\sigma(j)},\mathbf{m}^{gt}_j)\\
&+\lambda_{chg}\mathcal{L}_{bce}(z_{\hat\sigma(j)},y_j)\Big],
\end{split}
\label{eq:set}
\end{equation}
where unmatched queries are trained towards $\nul$ at both dates with weight $0.1$, and $\mathcal{L}_{mask}$ uses importance-sampled points (oversampling ratio $3$, importance ratio $0.75$). The set loss is applied to the prediction before the first decoder layer ($k=0$) and to the output of each layer ($k=1,2,3$). The dense auxiliary losses comprise the mean cross-entropy of the two semantic heads, $\mathcal{L}_{seg}$; a binary cross-entropy, in which pixels within three pixels of a ground-truth change boundary are up-weighted by a factor of four, plus a dice loss of the change head on $\mP_c$, $\mathcal{L}_{c}$; and the deep change supervision of the pyramid, $\mathcal{L}_{ms}$, averaged over the four levels. The complete objective is
\begin{equation}
\mathcal{L}=\sum_{k=0}^{3}\mathcal{L}^{(k)}_{set}+\lambda_{seg}\mathcal{L}_{seg}+\lambda_{c}\mathcal{L}_{c}+\lambda_{ms}\mathcal{L}_{ms},
\label{eq:total}
\end{equation}
with $\lambda_{cls}=0.2$, $\lambda_m=0.5$ for both the BCE and the dice term, $\lambda_{chg}=0.1$ and $\lambda_{seg}=\lambda_c=\lambda_{ms}=1$. The set losses are summed over four decoder outputs, and their weights, ten times lower than those of Mask2Former~\cite{cheng2022mask2former}, balance them against the dense change supervision of the shared encoder and pyramid.

\section{Experimental Results}
\label{sec:exp}

\subsection{Dataset Description}
We evaluate on three SCD benchmarks (Table~\ref{tab:datasets}). Their spatial resolution ranges from 0.1 to 30\,m: SECOND contains very-high-resolution aerial imagery at 0.5 to 3\,m, Landsat-SCD medium-resolution Landsat imagery at 30\,m, where object boundaries fall inside mixed pixels, and Hi-UCD ultra-high-resolution aerial imagery at 0.1\,m, where individual roofs, roads and greenhouses are resolved. Their label spaces have 7, 5 and 10 classes, which corresponds to 25 to 100 possible class pairs. SECOND and Landsat-SCD label land cover only inside changed regions, whereas Hi-UCD provides full-coverage land-cover maps at both dates. The scenes cover three Chinese cities, an arid agricultural region in Xinjiang and the city of Tallinn.

\subsubsection{SECOND} The SEmantic Change detectiON Dataset~\cite{yang2021asymmetric} contains $4\,662$ aerial image pairs of $512\times512$ pixels, collected from several platforms and sensors over Hangzhou, Chengdu and Shanghai, with six land-cover classes (non-vegetated ground, tree, low vegetation, water, building and playground) annotated inside changed regions, which form 30 change types. We use the official split of $2\,968$ training and $1\,694$ test pairs and hold out 10\,\% of the training pairs ($296$) for validation. SECOND is the de facto standard benchmark for SCD.
\subsubsection{Landsat-SCD} This dataset~\cite{yuan2022landsatscd} covers Tumushuke, Xinjiang, China, with Landsat imagery from 1990 to 2020 at 30\,m resolution and four classes (farmland, desert, building and water) annotated inside changed regions. Most of its released pairs are flipped, masked or resized copies of the original images; we use the $2\,385$ original pairs of $416\times416$ pixels and the public training, validation and test lists of $1\,431$/$477$/$477$ pairs.
\subsubsection{Hi-UCD} We use the mini release of Hi-UCD~\cite{tian2022hiucd}, which provides 0.1\,m aerial imagery of Tallinn, Estonia, with nine full-coverage classes (water, grassland, woodland, bare land, building, greenhouse, road, bridge and others). Training pairs were acquired in 2017 and 2018 and test pairs in 2018 and 2019; the $1\,024\times1\,024$ tiles of the training, validation and test sets are cropped into $256\times256$ tiles ($4\,800$/$944$/$6\,176$ pairs). Unlabeled pixels are treated as a class of their own in training and evaluation.

\begin{table}[t]
\centering
\caption{Datasets used in the experiments. $K$ includes the no-change class (``changed'' annotation) or the unlabeled class (``full'' annotation). The SECOND validation pairs are held out from its official training set.}
\label{tab:datasets}
\small
\setlength{\tabcolsep}{6pt}
\begin{tabular}{lccccc}
\toprule
Dataset & Sensor, resolution & Train / Val / Test & Size & $K$ & Labels \\
\midrule
SECOND & aerial, 0.5 to 3\,m & 2\,672 / 296 / 1\,694 & 512 & 7 & changed \\
Landsat-SCD & Landsat, 30\,m & 1\,431 / 477 / 477 & 416 & 5 & changed \\
Hi-UCD & aerial, 0.1\,m & 4\,800 / 944 / 6\,176 & 256 & 10 & full \\
\bottomrule
\end{tabular}
\end{table}

\subsection{Baseline Approaches}
MaskSCD is compared with fourteen models of thirteen SCD methods that represent the main design families of recent SCD research. The CNN multi-task networks are SCanNet~\cite{ding2024scannet}, trained with its pseudo-label scheme, ClearSCD~\cite{tang2024clearscd}, LSAFNet~\cite{zhou2024lsafnet} and TripleS~\cite{tan2025triples}. ClearSCD and TripleS build on EfficientNet-B0 encoders and use the semantic posteriors of the two dates in their change heads, and TripleS trains its network with stepwise multi-task optimization and selective parameter binding. HGINet~\cite{long2024hginet} and CPGNet~\cite{long2025cpgnet} use PVTv2 encoders, with graph-based semantic interaction and change-prior guidance, respectively. The state-space models are ChangeMamba~\cite{chen2024changemamba}, GSTM-SCD~\cite{liu2025gstmscd}, FoBa~\cite{zhang2025foba} and Mamba-FCS~\cite{wijenayake2025mambafcs}. GSTM-SCD combines a tree-topology state-space encoder with graph-based spatio-temporal modeling, and the other three use the VMamba-B encoder~\cite{liu2024vmamba}. PerASCD~\cite{shen2026perascd} is evaluated with a VMamba-B encoder (PerASCD$_\mathrm{VM}$) and with the PerA foundation model of about 500\,M parameters (PerASCD$_\mathrm{PerA}$), and SemDINO~\cite{tong2026semdino} uses frozen DINOv3 ViT-L features~\cite{simeoni2025dinov3}. DMCTNet~\cite{yang2025dmctnet}, the only other SCD network with a query-based mask decoder, segments each date with its own queries on the features of a lightweight Segment Anything encoder~\cite{kirillov2023sam} and gates the two semantic maps with a per-pixel change map.

\subsection{Implementation Details}
\label{sec:impl}
All methods are implemented in PyTorch and trained in one code base under a common protocol. They share the data splits, the flip and rotation augmentations (random horizontal and vertical flips and $90^\circ$ rotations), the number of epochs (50 on SECOND, 120 on Landsat-SCD and 100 on Hi-UCD), the batch size (8 for $416$- and $512$-pixel inputs and 16 for $256$-pixel inputs), the checkpoint selection and the test-time augmentation, and they keep their own losses, optimizers and learning rates. On Hi-UCD, the baselines receive the land-cover maps of both dates and the change map derived from them, and each baseline predicts the change map with its own change branch. Each encoder is initialized with the pre-trained weights used by its original method. On SECOND, all methods use an exponential moving average of the weights with decay $0.999$. For each model and dataset, the checkpoint with the best validation \SeK\ is selected, and the test predictions are averaged over the eight dihedral transforms. Each method and each ablation variant is trained twice on each dataset, and all results are the mean of the two training runs.

MaskSCD uses one configuration on all three datasets. It is trained with AdamW with a learning rate of $10^{-4}$ for all parameters and a weight decay of $0.05$, and with a stochastic depth rate of $0.1$ in the encoder. The schedule consists of $500$ warm-up iterations followed by polynomial decay with power $0.9$, with gradient clipping at $1.0$ and bfloat16 mixed precision. Each model is trained on a single NVIDIA GH200 GPU.

\subsection{Evaluation Metrics}
Following~\cite{yang2021asymmetric,ding2022bisrnet}, the standard SCD metrics are computed from the hybrid confusion matrix $Q$ accumulated over both dates, whose labels are $0$ for unchanged pixels and the land-cover class for changed pixels. We report the overall accuracy (OA), the mean intersection over union of the changed and unchanged regions (mIoU), the separated kappa coefficient
\begin{equation}
\SeK=e^{\mathrm{IoU}^c-1}\cdot\frac{\hat\rho-\hat\eta}{1-\hat\eta},
\label{eq:sek}
\end{equation}
where $\hat\rho$ and $\hat\eta$ are the observed and the expected agreement of $Q$ after its entry for pixels that are unchanged in both the prediction and the ground truth is set to zero and IoU$^c$ is the IoU of the changed region, and the semantic change $F_1$ score \Fscd, the harmonic mean of the precision and recall of ``changed with the correct class''.

A transition-level metric complements them. The transition $F_1$ score is $F_T=2\,n_{ok}/(n_{pred}+n_{gt})$, where $n_{ok}$ counts the pixels whose predicted transition is entirely correct, \ie, detected as changed with both classes right, and $n_{pred}$ and $n_{gt}$ the predicted and ground-truth changed pixels. Unlike \Fscd, which scores the semantic labels of the two dates separately, $F_T$ credits a changed pixel only if its complete ``from--to'' statement is correct. As a pixel-weighted score, it counts common and rare transitions in proportion to their area. Finally, the contradictory change pixels (CCP) measure the share of a model's predicted changed pixels whose change decision contradicts its own semantic maps: pixels predicted as changed but assigned the same class at both dates, or a no-change class at either date on datasets of the ``changed'' type. Lower is better. A model that derives the change map from the class pair cannot produce the first kind of contradiction.

\subsection{Comparison With State-of-the-Art Methods}

\begin{table*}[t]
\centering
\caption{Quantitative comparison on the three datasets (\%). The best and second-best results are highlighted in \textcolor{red}{\textbf{red}} and \textcolor{blue}{blue}, respectively.}
\label{tab:main}
\setlength{\tabcolsep}{1pt}
\resizebox{\textwidth}{!}{\fontsize{7}{8.5}\selectfont\begin{tabularx}{\textwidth}{@{}c@{\hspace{3pt}}l*{15}{Y}@{}}
\toprule
 & & SCanNet & ClearSCD & LSAFNet & TripleS & HGINet & CPGNet & Change-\newline Mamba & GSTM-\newline SCD & FoBa & Mamba-\newline FCS & PerASCD\newline VM & PerASCD\newline PerA & SemDINO & DMCTNet & MaskSCD \\
\midrule
\multirow{5}{*}{\rotatebox[origin=c]{90}{SECOND}} & OA & 88.42 & 86.83 & 88.39 & 85.74 & 88.18 & 88.42 & 88.92 & 87.44 & 89.08 & \textcolor{blue}{89.27} & 87.48 & 88.86 & 88.69 & 87.62 & \textcolor{red}{\textbf{89.32}} \\
 & mIoU & 73.86 & 70.37 & 73.79 & 69.39 & 73.02 & 73.60 & 74.17 & 73.05 & 74.33 & \textcolor{blue}{74.77} & 72.07 & 74.37 & 73.86 & 73.88 & \textcolor{red}{\textbf{74.98}} \\
 & SeK & 24.92 & 18.46 & 24.79 & 15.98 & 23.64 & 24.54 & 25.18 & 24.32 & 25.92 & \textcolor{blue}{26.72} & 21.66 & 26.15 & 24.95 & 25.54 & \textcolor{red}{\textbf{27.04}} \\
 & $F_{scd}$ & 65.24 & 59.26 & 65.16 & 56.42 & 64.33 & 64.94 & 65.25 & 64.53 & 66.27 & \textcolor{blue}{66.92} & 62.38 & 66.50 & 65.38 & 64.88 & \textcolor{red}{\textbf{67.11}} \\
 & $F_T$ & 57.50 & 50.51 & 57.37 & 46.66 & 56.80 & 57.28 & 57.30 & 56.61 & 59.02 & \textcolor{blue}{59.75} & 54.16 & 59.22 & 57.73 & 56.56 & \textcolor{red}{\textbf{59.94}} \\
\midrule
\multirow{5}{*}{\rotatebox[origin=c]{90}{\shortstack{Landsat-\\SCD}}} & OA & 97.45 & 93.00 & \textcolor{blue}{97.56} & 91.04 & 95.18 & 96.18 & 96.15 & 94.93 & 96.32 & 96.48 & 95.19 & 93.61 & 96.84 & 96.88 & \textcolor{red}{\textbf{98.04}} \\
 & mIoU & 92.29 & 80.68 & \textcolor{blue}{92.61} & 77.72 & 86.52 & 89.04 & 88.90 & 86.32 & 89.03 & 89.37 & 86.39 & 82.83 & 90.58 & 91.07 & \textcolor{red}{\textbf{93.91}} \\
 & SeK & 70.81 & 38.49 & \textcolor{blue}{71.85} & 31.30 & 52.75 & 60.30 & 59.63 & 53.05 & 60.80 & 61.99 & 53.20 & 43.92 & 65.53 & 66.62 & \textcolor{red}{\textbf{76.22}} \\
 & $F_{scd}$ & 92.72 & 78.75 & \textcolor{blue}{93.02} & 73.57 & 85.76 & 88.89 & 88.58 & 85.87 & 89.40 & 89.95 & 86.18 & 81.61 & 91.07 & 91.13 & \textcolor{red}{\textbf{94.34}} \\
 & $F_T$ & 92.16 & 76.22 & \textcolor{blue}{92.54} & 70.01 & 84.08 & 87.68 & 87.17 & 84.25 & 88.70 & 89.43 & 85.10 & 80.02 & 90.48 & 90.38 & \textcolor{red}{\textbf{93.93}} \\
\midrule
\multirow{5}{*}{\rotatebox[origin=c]{90}{Hi-UCD}} & OA & 91.46 & 94.44 & 94.68 & 93.47 & 94.59 & 95.13 & 95.13 & 94.38 & 94.69 & \textcolor{blue}{95.39} & 92.28 & 94.25 & 94.60 & 91.57 & \textcolor{red}{\textbf{95.72}} \\
 & mIoU & 66.53 & 71.17 & 72.24 & 69.76 & \textcolor{blue}{75.51} & \textcolor{red}{\textbf{76.12}} & 73.20 & 71.11 & 69.57 & 73.58 & 66.84 & 71.43 & 70.02 & 67.68 & 75.27 \\
 & SeK & 6.68 & 13.27 & 15.19 & 9.48 & 20.61 & \textcolor{blue}{22.83} & 17.88 & 11.66 & 12.99 & 19.99 & 6.00 & 13.80 & 12.62 & 8.87 & \textcolor{red}{\textbf{23.02}} \\
 & $F_{scd}$ & 38.07 & 47.85 & 50.72 & 42.86 & 54.93 & \textcolor{blue}{58.30} & 53.55 & 46.15 & 49.83 & 57.77 & 38.20 & 49.53 & 48.21 & 40.50 & \textcolor{red}{\textbf{60.49}} \\
 & $F_T$ & 26.79 & 36.28 & 41.64 & 31.44 & 43.54 & 47.07 & 42.88 & 32.44 & 42.07 & \textcolor{blue}{49.89} & 26.97 & 40.25 & 39.28 & 28.57 & \textcolor{red}{\textbf{52.39}} \\
\bottomrule
\end{tabularx}}
\end{table*}

\begin{table*}[t]
\centering
\caption{Model complexity and mean accuracy over the three datasets. Complexity refers to one $512\times512$ pair on one GH200 GPU in bfloat16 without test-time augmentation; parameters include frozen encoders, and one multiply-accumulate counts as one FLOP. SeK (F) keeps only the predicted changed pixels whose two classes form a valid transition, and SeK (P) derives the change map from the two semantic maps (post-classification comparison). The best and second-best results are highlighted in \textcolor{red}{\textbf{red}} and \textcolor{blue}{blue}, respectively.}
\label{tab:rank}
\scriptsize
\setlength{\tabcolsep}{2pt}
\begin{tabularx}{\textwidth}{@{}l*{9}{Y}@{}}
\toprule
\multirow{2}{*}{Method} & \multicolumn{4}{c}{Complexity} & \multicolumn{5}{c}{Mean accuracy over the three datasets (\%)} \\
\cmidrule(lr){2-5} \cmidrule(l){6-10}
 & Params (M) & GFLOPs & Time (ms) & Memory (GB) & SeK & $F_{scd}$ & $F_T$ & SeK (F) & SeK (P) \\
\midrule
SCanNet & 27.9 & 265.2 & 18.4 & 0.49 & 34.13 & 65.34 & 58.82 & 34.91 & \textcolor{blue}{32.04} \\
ClearSCD & 5.8 & 20.0 & 28.0 & 0.14 & 23.41 & 61.95 & 54.34 & 23.96 & 11.25 \\
LSAFNet & 28.4 & 520.3 & 26.3 & 0.45 & \textcolor{blue}{37.28} & 69.63 & 63.85 & \textcolor{blue}{37.48} & 29.44 \\
TripleS & 8.6 & 30.3 & 33.4 & 0.16 & 18.92 & 57.62 & 49.37 & 19.56 & 8.02 \\
HGINet & 27.7 & 58.2 & 43.4 & 0.26 & 32.33 & 68.34 & 61.47 & 32.49 & 13.39 \\
CPGNet & 55.4 & 86.9 & 67.6 & 0.43 & 35.89 & 70.71 & 64.01 & 36.35 & 24.50 \\
ChangeMamba & 97.0 & 229.0 & 79.5 & 1.11 & 34.23 & 69.13 & 62.45 & 34.69 & 21.25 \\
GSTM-SCD & 41.9 & 66.8 & 77.7 & 0.54 & 29.68 & 65.51 & 57.77 & 31.49 & 24.21 \\
FoBa & 93.6 & 179.2 & 89.5 & 0.84 & 33.23 & 68.50 & 63.26 & 33.27 & 17.69 \\
Mamba-FCS & 206.1 & 280.7 & 115.1 & 1.47 & 36.24 & \textcolor{blue}{71.54} & \textcolor{blue}{66.35} & 36.26 & 16.05 \\
PerASCD$_\mathrm{VM}$ & 95.1 & 213.2 & 94.6 & 0.61 & 26.95 & 62.25 & 55.41 & 27.24 & 19.35 \\
PerASCD$_\mathrm{PerA}$ & 549.6 & 1146.0 & 123.3 & 5.20 & 27.96 & 65.88 & 59.83 & 28.24 & 20.69 \\
SemDINO & 340.4 & 976.7 & 89.5 & 1.61 & 34.37 & 68.22 & 62.50 & 34.64 & 28.59 \\
DMCTNet & 48.1 & 964.7 & 84.3 & 1.48 & 33.68 & 65.51 & 58.50 & 34.08 & 8.88 \\
MaskSCD & 106.6 & 379.0 & 55.1 & 0.98 & \textcolor{red}{\textbf{42.09}} & \textcolor{red}{\textbf{73.98}} & \textcolor{red}{\textbf{68.75}} & \textcolor{red}{\textbf{42.09}} & \textcolor{red}{\textbf{42.09}} \\
\bottomrule
\end{tabularx}
\end{table*}

\begin{table}[t]
\centering
\caption{Contradictory change pixels (CCP, lower is better) on the three datasets (\%). The best and second-best results are highlighted in \textcolor{red}{\textbf{red}} and \textcolor{blue}{blue}, respectively.}
\label{tab:ccp}
\scriptsize
\setlength{\tabcolsep}{4pt}
\begin{tabular}{lccc}
\toprule
Method & SECOND & Landsat-SCD & Hi-UCD \\
\midrule
SCanNet & 0.41 & 1.30 & 28.97 \\
ClearSCD & 5.49 & 6.13 & 5.90 \\
LSAFNet & 0.27 & 0.14 & 14.97 \\
TripleS & 10.09 & 8.17 & 11.80 \\
HGINet & 1.47 & 1.95 & 5.40 \\
CPGNet & 2.20 & 2.18 & 4.46 \\
ChangeMamba & 2.35 & 3.20 & 5.36 \\
GSTM-SCD & 4.31 & 6.33 & 17.18 \\
FoBa & \textcolor{blue}{0.12} & \textcolor{blue}{0.08} & 5.03 \\
Mamba-FCS & 0.30 & 0.25 & \textcolor{blue}{3.48} \\
PerASCD$_\mathrm{VM}$ & 0.70 & 0.43 & 27.49 \\
PerASCD$_\mathrm{PerA}$ & 0.22 & 0.51 & 17.83 \\
SemDINO & 0.43 & 0.68 & 12.81 \\
DMCTNet & 1.09 & 0.11 & 29.82 \\
MaskSCD & \textcolor{red}{\textbf{0.00}} & \textcolor{red}{\textbf{0.00}} & \textcolor{red}{\textbf{0.00}} \\
\bottomrule
\end{tabular}
\end{table}

\begin{figure*}[t]
\centering
\includegraphics[width=\textwidth]{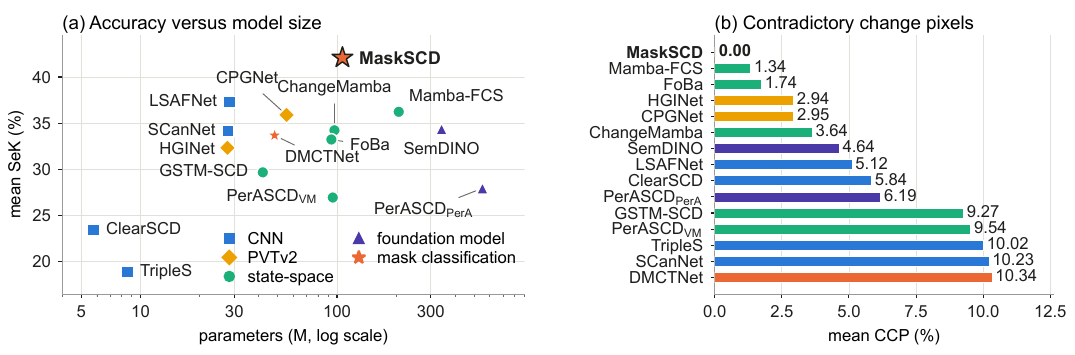}
\caption{Overall comparison on the three datasets. (a) Mean \SeK\ versus number of parameters. (b) Mean CCP.}
\label{fig:charts}
\end{figure*}

\begin{figure*}[t]
\centering
\includegraphics[width=\textwidth]{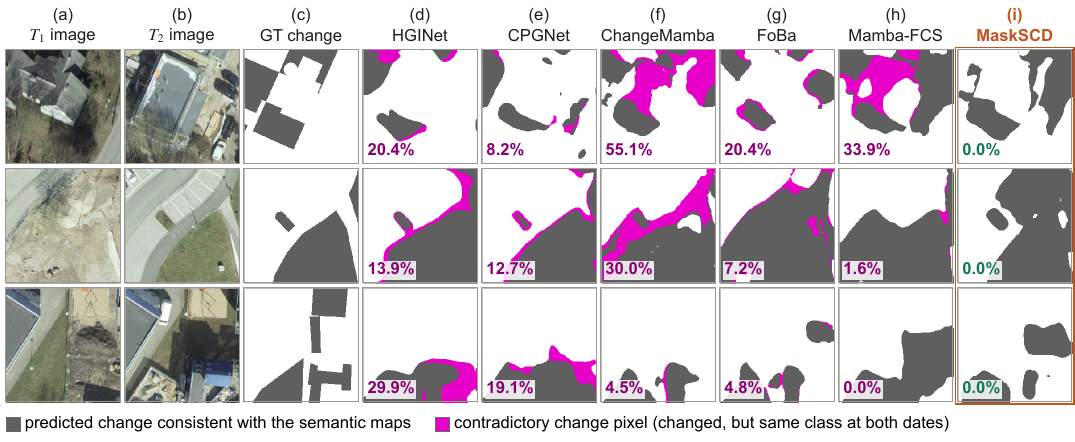}
\caption{Contradictory change pixels (magenta) on Hi-UCD. Predicted change is shown in gray; the number in each panel is the share of contradictory pixels in the predicted change.}
\label{fig:consist}
\end{figure*}

\begin{figure*}[t]
\centering
\includegraphics[width=\textwidth]{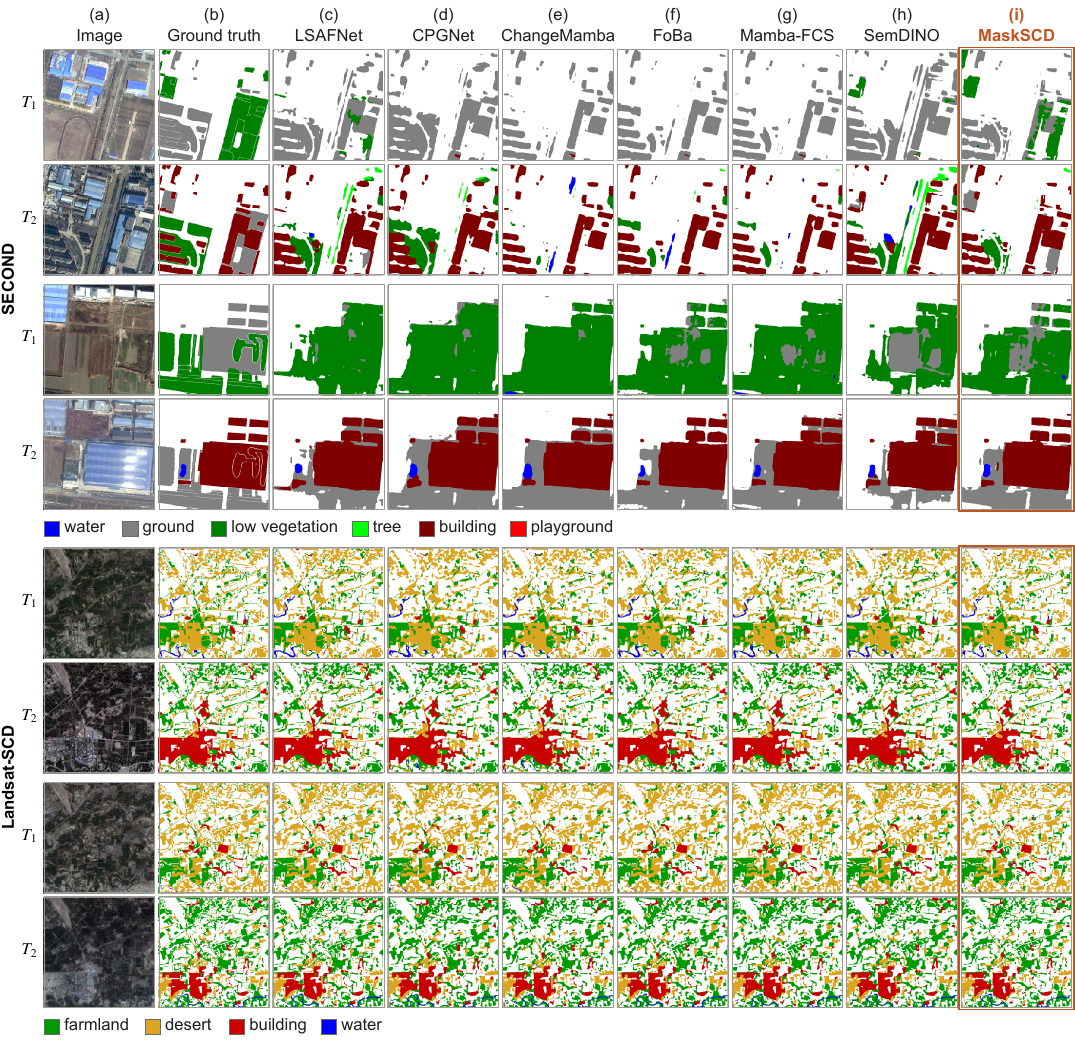}
\caption{Semantic change maps on SECOND (top) and Landsat-SCD (bottom), with two rows ($T_1$, $T_2$) per sample. Unchanged pixels are white.}
\label{fig:qual1}
\end{figure*}

\begin{figure*}[t]
\centering
\includegraphics[width=\textwidth]{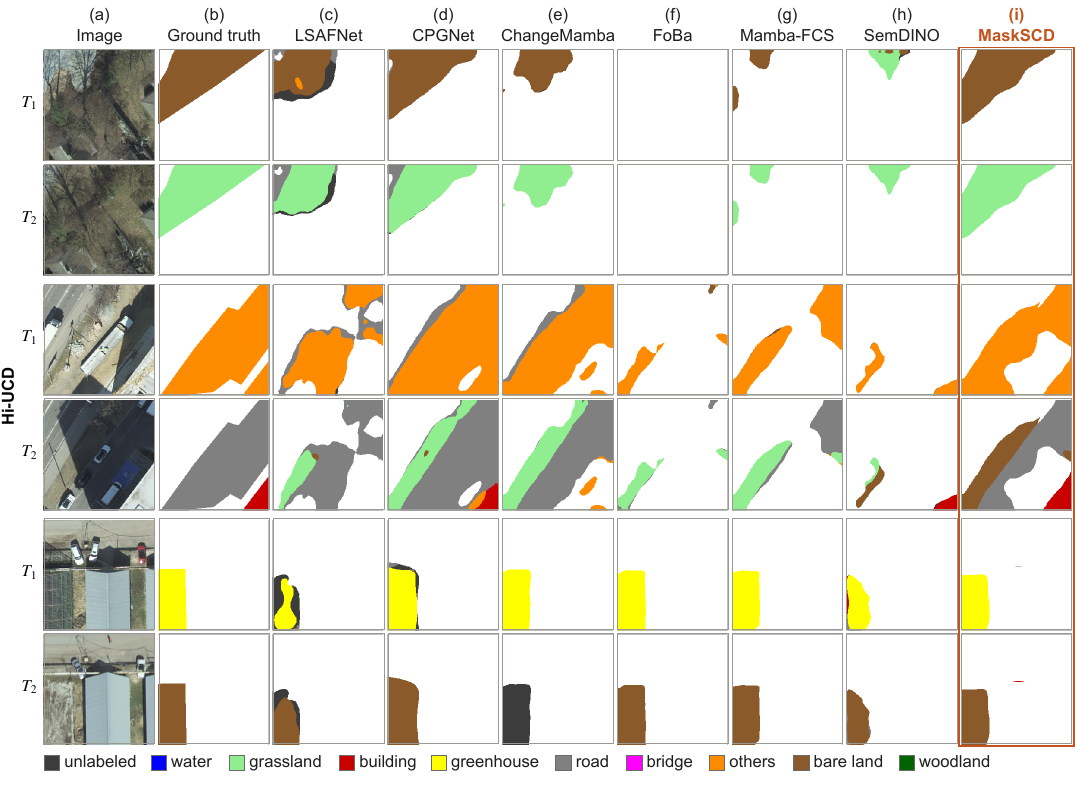}
\caption{Semantic change maps on Hi-UCD, with two rows ($T_1$, $T_2$) per sample. Unchanged pixels are white.}
\label{fig:qual2}
\end{figure*}

\begin{figure*}[t]
\centering
\includegraphics[width=\textwidth]{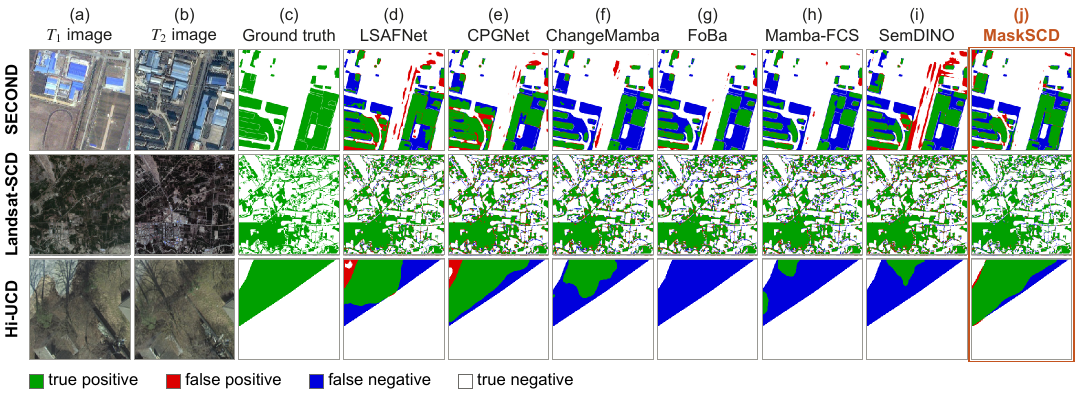}
\caption{Binary change maps on the three datasets: true positives (green), false positives (red), false negatives (blue) and true negatives (white).}
\label{fig:err}
\end{figure*}

\subsubsection{Quantitative Comparisons}
Table~\ref{tab:main} reports OA, mIoU, \SeK, \Fscd\ and $F_T$ of all methods on the three datasets, and Table~\ref{tab:rank} gives the model complexity and the mean accuracy over the three datasets. MaskSCD achieves the highest mean \SeK\ and \Fscd, with a clear margin over the second-best methods, LSAFNet in \SeK\ and Mamba-FCS in \Fscd, and it ranks first in \SeK, \Fscd\ and $F_T$ on every dataset. The second-best method, in contrast, differs between datasets. Mamba-FCS is second on SECOND but falls behind on Landsat-SCD, LSAFNet is second on Landsat-SCD but only mid-field on SECOND, and CPGNet is second on Hi-UCD but in the lower half on SECOND.

On SECOND, the most widely used benchmark, MaskSCD is the best method on all five metrics. It outperforms its closest competitors, Mamba-FCS and the foundation-model-based PerASCD$_\mathrm{PerA}$, with far fewer parameters than the latter. On Landsat-SCD, MaskSCD is again the best on all five metrics. The CNN multi-task networks SCanNet and LSAFNet are strong on this medium-resolution dataset, and MaskSCD still surpasses both, while all four state-space models fall well behind. On the fine-grained Hi-UCD, MaskSCD achieves the best \SeK, \Fscd, $F_T$ and OA, closely followed by CPGNet in \SeK, whereas CPGNet and HGINet obtain a higher mIoU. ClearSCD and TripleS are the two weakest methods on SECOND and Landsat-SCD, where land cover is annotated only inside changed regions. The query-based DMCTNet ranks in the middle of the field in mean \SeK\ over the three datasets, and MaskSCD surpasses it on every dataset and metric.

Fig.~\ref{fig:charts}(a) relates accuracy to model size. MaskSCD lies above all other methods, including Mamba-FCS, SemDINO and PerASCD$_\mathrm{PerA}$, which have two to five times as many parameters. At inference, it is faster than all state-space, foundation-model and query-based baselines (Table~\ref{tab:rank}). The cross-temporal fusion and the change pyramid of the \BFP\ add about 7\,M parameters.

\subsubsection{Transition $F_1$ and Consistency}
As shown in Table~\ref{tab:main}, MaskSCD obtains the highest transition $F_1$ on all three datasets. The margin over the second-best model is largest on the fine-grained Hi-UCD. There, CPGNet localizes change with a higher mIoU, but MaskSCD exceeds it by a wide margin in $F_T$, so a larger share of its detected change carries the correct class pair.

Table~\ref{tab:ccp} and Fig.~\ref{fig:charts}(b) show the benefit of deriving change from the class pair. Every multi-task network produces contradictory change pixels. On the two datasets that are annotated inside changed regions only, their share is mostly small, but it rises sharply on the full-coverage Hi-UCD, where many baselines declare change over regions that they simultaneously label with the same class at both dates. Mask decoding alone does not remove them, as DMCTNet has the highest share of contradictory change pixels of all methods on Hi-UCD. MaskSCD produces no contradictory change pixels on any of the three datasets. Its pair aggregation cannot assign the same class to both dates of a changed pixel, and on SECOND and Landsat-SCD none of its changed pixels pairs the no-change class with a land-cover class, so every changed pixel it outputs comes with a valid transition.

Consistency can also be imposed on a multi-task network after inference, by keeping only the changed pixels whose two classes form a valid transition (F in Table~\ref{tab:rank}) or by taking the change map from the two semantic maps as in post-classification comparison (P). The first changes the mean \SeK\ of the baselines only slightly, and the second lowers it, for most baselines sharply, as the classification errors of both dates accumulate in the change map. With either read-out, the baselines remain well below MaskSCD in mean \SeK.

Fig.~\ref{fig:consist} visualizes these contradictions on Hi-UCD. The magenta pixels of the multi-task models concentrate along object boundaries and cover entire objects where the change and semantic decoders disagree, in one scene more than half of the predicted change.

\subsubsection{Qualitative Comparisons}
Figs.~\ref{fig:qual1} and \ref{fig:qual2} show the semantic change maps of the compared methods on the three datasets, and Fig.~\ref{fig:err} shows the corresponding binary change maps. From these representative results, MaskSCD exhibits the following advantages.
\begin{itemize}
\item Object integrity: Changed objects such as buildings and agricultural parcels are delineated as complete regions with sharp boundaries (\eg, the new buildings in the first SECOND sample of Fig.~\ref{fig:qual1} and the bare land converted to grassland in the first Hi-UCD sample of Fig.~\ref{fig:qual2}), whereas per-pixel networks fragment them into several blobs, leave holes inside homogeneous objects or blur the boundaries between adjacent transitions.
\item Consistent transitions: In the results of the multi-task networks, the $T_1$ and $T_2$ maps of the same object often disagree in extent, or a region is marked as changed in one map and not in the other, which corresponds to contradictory change pixels. In the MaskSCD results, the two maps always cover the same changed area and differ only in color, \ie, in class.
\item Fewer pseudo-changes and correct classes: As both temporal states of a query read the same masked region, MaskSCD produces fewer false alarms in unchanged built-up and vegetated areas (Fig.~\ref{fig:err}), which appear as red false-positive regions in the maps of the baselines. It also assigns the correct class to detected change more often, as reflected by its transition $F_1$ on all three datasets.
\end{itemize}

\subsection{Ablation Studies}
\label{sec:ablation}

\begin{table}[t]
\centering
\caption{Ablation of the bitemporal masked-attention decoder, the bitemporal fusion pyramid and the backbone in terms of the mean mIoU, \SeK, \Fscd\ and $F_T$ over the three datasets (\%). Each variant changes one setting of MaskSCD (\#01).}
\label{tab:ablation}
\footnotesize
\setlength{\tabcolsep}{2.5pt}
\begin{tabularx}{\columnwidth}{c|>{\centering\arraybackslash}X|cccc}
\toprule
\multirow{2}{*}{No.} & \multirow{2}{*}{Variants} & \multicolumn{4}{c}{Metrics} \\ \cmidrule{3-6}
 & & mIoU & SeK & $F_{scd}$ & $F_T$ \\
\midrule
\#01 & MaskSCD & 81.39 & 42.09 & 73.98 & 68.75 \\
\midrule
\multicolumn{6}{>{\raggedright\arraybackslash}p{\dimexpr\linewidth-2\tabcolsep\relax}}{\textit{Decoder components}} \\
\midrule
\#02 & w/o masked attention & 80.51 & 40.35 & 71.91 & 66.51 \\
\#03 & w/o joint self-attention & 80.52 & 40.75 & 73.00 & 68.02 \\
\#04 & w/o masked attention and joint self-attention & 79.99 & 39.21 & 70.92 & 65.24 \\
\midrule
\multicolumn{6}{>{\raggedright\arraybackslash}p{\dimexpr\linewidth-2\tabcolsep\relax}}{\textit{Number of queries} (default: $N=100$)} \\
\midrule
\#05 & $N=25$ & 81.08 & 41.66 & 73.59 & 68.43 \\
\#06 & $N=50$ & 81.30 & 41.98 & 73.92 & 68.72 \\
\#07 & $N=150$ & 81.25 & 41.80 & 73.63 & 68.46 \\
\midrule
\multicolumn{6}{>{\raggedright\arraybackslash}p{\dimexpr\linewidth-2\tabcolsep\relax}}{\textit{Bitemporal fusion pyramid}} \\
\midrule
\#08 & w/o cross-temporal fusion & 81.14 & 41.66 & 73.62 & 68.31 \\
\midrule
\multicolumn{6}{>{\raggedright\arraybackslash}p{\dimexpr\linewidth-2\tabcolsep\relax}}{\textit{Backbone} (default: Swin-B)} \\
\midrule
\#09 & Swin-T & 79.99 & 38.63 & 72.18 & 67.11 \\
\#10 & Swin-S & 80.82 & 40.50 & 73.21 & 67.94 \\
\#11 & Swin-L & 81.50 & 42.69 & 74.15 & 68.93 \\
\bottomrule
\end{tabularx}
\end{table}

\begin{figure*}[t]
\centering
\includegraphics[width=\textwidth]{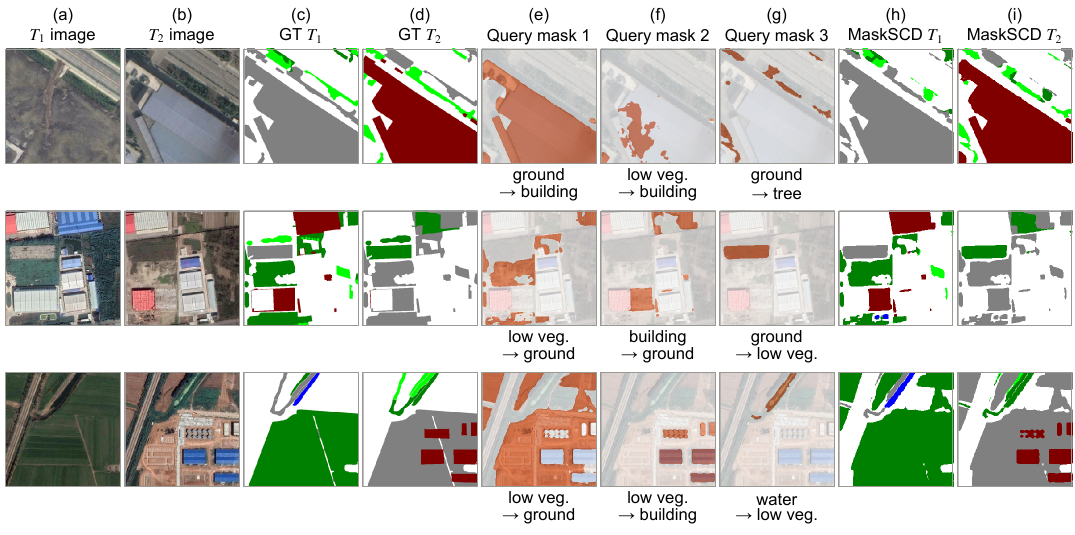}
\caption{Formation of the predicted masks on SECOND. (a), (b) Input images. (c), (d) Ground truth. (e), (f), (g) The three largest query masks with a changed class pair and their predicted transition. (h), (i) Semantic change maps after pair aggregation.}
\label{fig:masks}
\end{figure*}

Table~\ref{tab:ablation} changes one setting of MaskSCD at a time. As in Table~\ref{tab:rank}, each value is the mean over the three datasets, and \#01 is the mean accuracy of MaskSCD.

\subsubsection{Effectiveness of the Bitemporal Masked-Attention Decoder}
\label{sec:design}
Without masked attention (\#02), each state attends to its whole image. All four metrics decrease, and the largest drops occur in \SeK, \Fscd\ and $F_T$, which score the class pairs of the changed pixels. Without the joint bitemporal self-attention (\#03), self-attention is applied to the states of each date separately and the two states of a query interact only through their shared mask, which lowers all four metrics. Removing both components (\#04) lowers \SeK\ by nearly three points, more than either removal alone.

$N=100$ queries give the best result, and fewer or more queries (\#05 to \#07) lower all four metrics.

\subsubsection{Effectiveness of the Bitemporal Fusion Pyramid}
Without the cross-temporal fusion (\#08), the two dates interact only through a temporal exchange, one transformer layer over the concatenated tokens of both dates at strides 32 and 16, and a $1\times1$ convolution of the per-date features and their absolute difference provides the change features of the pyramid. All four metrics decrease, so the cross-temporal fusion of the \BFP\ strengthens the bitemporal interaction.

\subsubsection{Choice of the Backbone}
Accuracy increases with the size of the Swin backbone (\#09 to \#11). Swin-L, with twice the parameters of Swin-B, raises the mean \SeK\ by 0.6 points, and Swin-B balances accuracy and cost.

\subsubsection{Formation of the Predicted Masks}
Fig.~\ref{fig:masks} shows how the final maps are composed. Individual queries cover coherent bitemporal segments, such as a factory hall built on former ground, a large field of low vegetation converted to bare ground or a narrow strip along a channel that changed from water to low vegetation, and each query carries one class for each date. Different transitions in the same scene are taken by different queries, and the pair aggregation of (\ref{eq:pair}) assembles them into semantic change maps whose changed extent is identical at both dates.

\FloatBarrier
\section{Conclusion}
\label{sec:conclusion}
This article has formulated SCD as the mask classification of the bitemporal label-pair image and has presented MaskSCD on this basis. MaskSCD predicts each bitemporal segment, a region of constant class pair, as one mask shared by both dates and one class per date, and derives the change map from the predicted class pair. A bitemporal masked-attention transformer decoder with two temporal states per query reads each segment in both images, and a bitemporal fusion pyramid provides its per-date memories and change-aware mask feature.

The experiments on SECOND, Landsat-SCD and Hi-UCD, which differ in sensor, resolution and number of land-cover classes, support four findings. First, the predicted class pair makes change and semantics consistent. All re-trained multi-task and query-based baselines produce contradictory change pixels, most of all on the full-coverage Hi-UCD, whereas MaskSCD produces none on any dataset. Enforcing consistency on the baselines after inference, by filtering invalid transitions or by post-classification comparison, does not reach the accuracy of MaskSCD. Second, predicting bitemporal segments with shared masks and class pairs improves accuracy. MaskSCD achieves the highest mean \SeK\ over the three datasets with a clear margin and ranks first in \SeK, \Fscd\ and transition $F_1$ on every dataset, with the largest transition $F_1$ margin on Hi-UCD, the dataset with the most classes. Third, the ablations support the design of the decoder and the fusion pyramid. Removing masked attention or the joint bitemporal self-attention lowers all four metrics, removing both lowers them the most, and the cross-temporal fusion of the \BFP\ strengthens the bitemporal interaction. Fourth, MaskSCD balances accuracy and inference cost. It is more accurate than baselines with several times as many parameters and runs faster than all state-space, foundation-model and query-based baselines.

Any hierarchical encoder can provide the four feature levels of MaskSCD, which opens the way to remote sensing foundation models as encoders. The shared-mask formulation also extends to image time series, such as the multitemporal Sentinel-2 mining footprints of EuroMineNet~\cite{yu2026euromininet}, where each segment would carry a land-cover trajectory over all acquisition dates. Trained models will be released in the code repository.

\bibliographystyle{IEEEtran}
\bibliography{refs}

\end{document}